\documentclass[sigconf]{acmart}

\usepackage[utf8]{inputenc}
\usepackage{graphicx}
\usepackage{array}
\usepackage{booktabs} 
\usepackage[export]{adjustbox} 
\usepackage{multirow}
\usepackage{subcaption} % 여러 그림 배치를 위한 패키지
\usepackage{tabularx}
\usepackage{float}      % [H]
\usepackage{placeins}
\usepackage{enumitem}
\usepackage{hyperref}% \FloatBarrier
\usepackage{ragged2e}
\usepackage{makecell}
\usepackage{caption}
\usepackage{ragged2e}
\DeclareCaptionLabelFormat{appendix_fig}{Figure A#2}
\usepackage[nameinlink]{cleveref}

\AtBeginDocument{%
  }

\copyrightyear{2026}
\acmYear{2026}
\setcopyright{cc}
\setcctype{by}
\acmDOI{}
\acmBooktitle{Proceedings of the HRI 2026 Workshop: Equitable Robotics for Wellbeing (Eq-RW) (HRI '26), March 16--19, 2026, Edinburgh, Scotland, United Kingdom}
\acmConference[HRI '26]{ACM/IEEE International Conference on Human-Robot Interaction (HRI 2026)}%
  {March 16--19, 2026}{Edinburgh, Scotland, United Kingdom}
\acmISBN{}

\begin{document}

%%
%% The "title" command has an optional parameter,
%% allowing the author to define a "short title" to be used in page headers.
\title{Towards Equitable Robotic Furnishing Agents for Aging-in-Place: ADL-Grounded Design Exploration}

%%
%% The "author" command and its associated commands are used to define
%% the authors and their affiliations.
%% Of note is the shared affiliation of the first two authors, and the
%% "authornote" and "authornotemark" commands
%% used to denote shared contribution to the research.

%\email{h.lee1@imperial.ac.uk}
%\email{hansoolee@kist.re.kr}

%\email{026110@kist.re.kr}

%\email{soobinpark@kist.re.kr}

%\email{sonakwak@kist.re.kr}

\author{Hansoo Lee}
\authornote{These authors are joint lead authors.} % <- 이게 첫 번째 note라 보통 * 로 표시됨
\affiliation{%
  \institution{Imperial College London, United Kingdom}
  \country{}
}
\affiliation{%
  \institution{KIST, Republic of Korea}
  \country{}
}

\author{Changhee Seo}
\authornotemark[1] % <- Hansoo의 equal-contrib note(*) 재사용
\affiliation{%
  \institution{KIST}
  \country{Republic of Korea}
}

\author{Subin Park}
\affiliation{%
  \institution{KIST}
  \country{Republic of Korea}
}

\author{Sonya S. Kwak}
\authornote{Corresponding author.} % <- 두 번째 note라 보통 † 로 표시됨
\affiliation{%
  \institution{KIST}
  \country{Republic of Korea}
}

\begin{comment}

\author{Hansoo Lee\thanks{These authors contributed equally.}}
\affiliation{%
  \institution{Imperial College London}
  \country{United Kingdom}
}
\affiliation{%
  \institution{KIST}
  \country{Republic of Korea}
}

\author{Changhee Seo\thanks{These authors contributed equally.}}
\affiliation{%
  \institution{KIST}
  \country{Republic of Korea}
}

\author{Subin Park}
\affiliation{%
  \institution{KIST}
  \country{Republic of Korea}
}

\author{Sonya S. Kwak\thanks{Corresponding author.}}
\affiliation{%
  \institution{KIST}
  \country{Republic of Korea}
}
\end{comment}

%%
%% By default, the full list of authors will be used in the page
%% headers. Often, this list is too long, and will overlap
%% other information printed in the page headers. This command allows
%% the author to define a more concise list
%% of authors' names for this purpose.
\renewcommand{\shortauthors}{lee et al.}

%%
%% The abstract is a short summary of the work to be presented in the
%% article.
\begin{abstract}
In aging-in-place contexts, small difficulties in Activities of Daily Living (ADL) can accumulate, affecting well-being through fatigue, anxiety, reduced autonomy, and safety risks. This position paper argues that robotics for older adult wellbeing must move beyond “convenience features” and centre equity, justice, and responsibility. We conducted ADL-grounded semi-structured interviews with four adults in their 70s–80s, identifying recurrent challenges (finding/\allowbreak organising items, taking medication, and transporting objects) and deriving requirements to reduce compounded cognitive–physical burden. Based on these insights, we propose an in-home robotic furnishing–agent concept leveraging computer vision and generative AI/\allowbreak LLMs for natural-language interaction, context-aware reminders, safe actuation, and user-centred transparency. We then report video-stimulated follow-up interviews with the same participants, highlighting preferences for confirmation before actuation, predictability, adjustable speed/\allowbreak autonomy, and multimodal feedback, as well as equity-related concerns. We conclude with open questions on evaluating and deploying equitable robotic wellbeing systems in real homes.
\end{abstract}

%%
%% The code below is generated by the tool at http://dl.acm.org/ccs.cfm.
%% Please copy and paste the code instead of the example below.
%%
% ACM CCS 2012


%%
%% Keywords. The author(s) should pick words that accurately describe
%% the work being presented. Separate the keywords with commas.
%% A "teaser" image appears between the author and affiliation
%% information and the body of the document, and typically spans the
%% page.

%%
%% This command processes the author and affiliation and title
%% information and builds the first part of the formatted document.
\maketitle

\section{Introduction}
In aging-in-place contexts, declines in everyday functional ability affect not only convenience but also safety, autonomy, and emotional well-being. To systematically understand such changes, healthcare and gerontology have long relied on standardised assessment frameworks such as the Activities of Daily Living (ADL) and Instrumental Activities of Daily Living (IADL). ADL indices have been widely used as representative measures to quantify functional status
in older adults~\cite{katz1963studies}, while IADL extends this perspective to more complex life tasks closely tied to independent living~\cite{lawton1969assessment}. A variety of instruments exist to evaluate ADL/IADL, and systematic reviews have discussed which measures may be appropriate for different aging-related research contexts.

Prior Human--Robot Interaction research has emphasised that assistive robots for independent living should be understood not as isolated functional tools, but as part of a broader ecology of everyday life~\cite{forlizzi2004assistive}. Related work also argues that assistive robots should be reframed beyond deficit compensation toward supporting successful aging, including older adults' autonomy and goal pursuit~\cite{lee2018reframing}. Design guidelines for robots supporting older adults further suggest that acceptance, perceived control, and trust in everyday contexts should be treated as core design variables alongside safety and usability~\cite{beer2012domesticated}.

Home interfaces are particularly shaped by factors that drive both acceptance and rejection. For example, research on voice interfaces for older adults reports that convenience must be considered together with concerns such as privacy, errors, and discomfort~\cite{portet2013design}. In addition, assistive systems suggest that as automation increases, attitudes toward user control and the willingness to delegate become central issues~\cite{viswanathan2014wizard}. To reflect the needs of diverse and potentially vulnerable users, participatory and co-design approaches can be effective; participatory work with older adults has demonstrated the value of mutual learning in shaping social robot design~\cite{lee2017steps}. More recently, landscape analyses and co-design toolkits for home-based elderly care robots have also highlighted the importance in the design process~\cite{bardaro2022robots}.

Meanwhile, distributed approaches such as robotic furniture and networked in-home assistive furniture suggest a path toward improving accessibility and sustainability by embedding capabilities into familiar furniture forms~\cite{de2016networked}. A growing body of research has explored the design and evaluation of robotic furnishings and modular smart furniture systems aimed at enabling independent living~\cite{verma2018design, merilampi2020modular, threatt2017design}.

Building on these discussions, we conducted ADL-grounded semi-structured interviews to identify recurrent pain points in older adults’ daily routines (finding/organising items, taking medication, and transporting objects)~\cite{katz1963studies,lawton1969assessment, jekel2015mild}. Based on these insights, we envisioned an equity-centred robotic furnishing-agent concept design that augments our existing robotic furnishing system with generative AI and computer vision to enhance context awareness, interaction, and assistive capabilities. We then illustrated this enhanced system through a set of representative usage scenarios. For evaluation, we presented participants with videos of the existing robotic furnishing in operation, accompanied by scenario explanations, in follow-up interviews with the same participants. From these sessions, we distilled key design requirements for equitable wellbeing support—such as confirmation before actuation, predictability, adjustable autonomy, and multimodal feedback. Synthesizing these empirical findings with prior work on delegation and trust in collaborative robotic systems~\cite{shin2023robot} and on an existing collaborative robot system~\cite{Kang2024CollaBot}, we argue that technological advancement should not simply mean stronger automation, but rather more responsible interaction that preserves safety and autonomy.

%Drawing on prior work on delegation and trust in collaborative robotic systems~\cite{shin2023robot} and on an existing collaborative robot system implementation~\cite{Kang2024CollaBot}, we argue that technological advancement should not simply mean stronger automation, but rather more responsible interaction that preserves safety and autonomy. 

Through this position paper, we aim to catalyse discussion on evaluating equitable robotic wellbeing systems in real homes, including issues of bias, safety, privacy, and sustainable deployment.
\FloatBarrier

\begin{table}[!htbp]
% \begin{table}[htbp]% 표 위치 고정 해제
\centering
\caption{Painpoints in older adults’ daily activities} 
\label{tab:painpoints} 
\vspace{-6pt}
\footnotesize
\setlength{\tabcolsep}{3pt} 
\renewcommand{\arraystretch}{1.2} 
\newcolumntype{L}[1]{>{\RaggedRight\arraybackslash}m{#1}}
\newcolumntype{C}[1]{>{\centering\arraybackslash}m{#1}}

\setlist[itemize]{
    leftmargin=1em, 
    nosep, 
    topsep=2pt, 
    before=\vspace{0.1\baselineskip}, 
    after=\vspace{-0.1\baselineskip}
}

\begin{tabular}{C{1.5cm} L{4.2cm} L{2.1cm}}
\toprule 
\textbf{Activity} & \centering \textbf{Participant statement} (P1-P4) & \centering \textbf{Painpoint} \tabularnewline 
\midrule 

Organizing \& Finding & 
\begin{itemize}
    \item Forget storage locations of in\-fre\-quent\-ly used items. (P1)
    \item Forget drawer contents despite category-based storage. (P3)
    \item Repeated drawer use while searching for items can cause un\-in\-tend\-ed col\-li\-sions. (P4)
    \item Laundry sorting triggers repeated opening/closing due to location lapses. (P3)
\end{itemize} &  Cognitive and physical strain from repeated drawer use
 \\ \addlinespace[0.8em]

Taking Medication & 
\begin{itemize}
    \item Miss medication when engaged in other activities. (P1)
    \item Keep empty snack items visible to confirm intake. (P2)
    \item Keep medication within sight/reach for timing-sensitive routines. (P3)
\end{itemize} & 
Mental load from remembering and checking medication intake \\ \addlinespace[0.8em]

Transporting Objects & 
\begin{itemize}
    \item Want help carrying heavy groceries. (P2)
    \item Laundry requires repeated carrying, which is exhausting. (P4)
    \item Difficulty moving large items through narrow spaces. (P2)
\end{itemize} & Physical fatigue from low strength and frequent repeated movement
 \\ 
\bottomrule 
\end{tabular}
\end{table}
%%===========================첫번째 테이블=================================

\section{ADL-Grounded Interviews with Older Adults}
\label{sec:needs_assessment}
To elicit pain points and functional requirements for older adult-centred robotic products, we conducted semi-structured interviews with four older adults (ages 70 to 90). The interview guide was structured around Activities of Daily Living (ADL) and Instrumental Activities of Daily Living (IADL), using concrete home routines to prompt recall of difficulties and coping strategies. We analysed statements to identify recurring activities and the underlying cognitive/\allowbreak physical constraints that shape them. Representative participant statements and the detailed pain points derived from them are summarised in Table~\ref{tab:painpoints}.

Across participants, three activities repeatedly surfaced as difficulties: (a) organising and finding items,
(b) taking medication, and (c) transporting objects. Importantly, these were rarely `purely physical' or
`purely cognitive'. For example, searching for items with combined memory demands (remembering what is stored where) with repeated drawer opening/closing, which can be tiring and can increase the risk of bumps or pinching. Medication routines introduced prospective-memory demands (remembering when and whether medication was taken). Transporting objects highlighted reduced strength and the burden of repeated trips. 

From these findings, we derive three design requirements that we frame as equity-relevant because they aim to reduce burden while preserving autonomy and safety: 

\begin{itemize}[leftmargin=1.4em, labelsep=0.5em, itemsep=0.2em, topsep=2pt, partopsep=0pt]
  \item \textbf{R1. Storage visibility and memory support} for recall-heavy organising/finding routines.
  \item \textbf{R2. Time- and context-aware medication reminders} that also support confirmation of intake.
  \item \textbf{R3. Physical assistance for repeated transport and re-organisation} tasks, with predictable
  behaviour and user control.
\end{itemize}

%\FloatBarrier

%%===========================첫번째 테이블=================================

\section{Concept: Conceptual Design of a Collaborative Robotic Furnishing–Agent}

\subsection{Baseline Collaborative Robotic Furnishing System}

This work builds on the CollaBot system developed in prior studies~\cite{Kang2024CollaBot, shin2023robot}. Current CollaBot system prototype prioritizes a limited yet reliable scope of automation, rather than a full autonomy level based on real agent. Specifically, it only includes (1) a cabinet whose drawers open and automatically close via button inputs on a smartphone app, (2) user-approach detection based on ROI-based skeleton tracking (rather than full-fledged activity recognition), and (3) a cart-like mobile platform that provides transport assistance by moving only to predefined locations (e.g., directly in front of the cabinet). This constrained implementation is advantageous for achieving stability and predictability in real home environments; however, it offers limited interaction channels and semantic understanding, making it insufficient to fully address the specific challenges identified in Section~\ref{sec:needs_assessment}, such as memory lapses in locating items, the prospective cognitive load of medication routines, and the physical fatigue associated with repetitive transport.

\subsection{Proposed Extension: Generative AI and Vision-Enhanced Interaction}
Recent advances in generative AI/LLMs and computer vision enable agents to interpret user intent and support more natural interaction. Building on this trend, we propose a conceptual extension that adds an LLM-augmented agent layer on top of the existing CollaBot capabilities. The core aim is not strong automation per sec, but rather to reduce user burden while preserving safety and autonomy by enabling (a) confirmable assistance, (b) explainable actions, and (c) adjustable autonomy.

\subsection{Scenarios}
\begin{figure}[!htbp]
    \centering
    \begin{subfigure}{\linewidth}
        \centering
        \includegraphics[width=\linewidth]{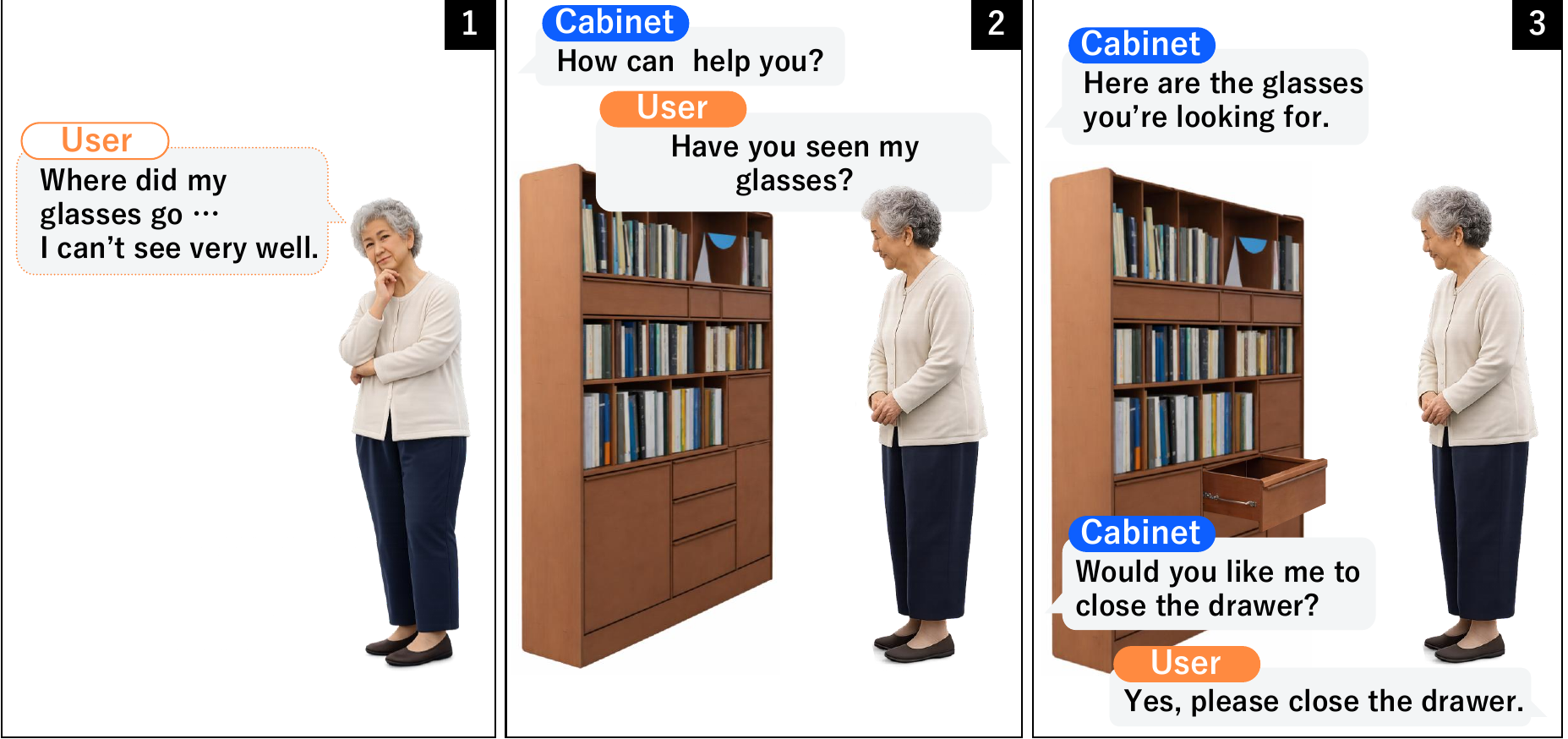}
        \caption{Finding an item scenario}
        \label{fig:app_1a}
        \Description{A storyboard showing a sequence of a robot helping a person find an item in a living room.}
    \end{subfigure}
    
    \begin{subfigure}{\linewidth}
        \centering
        \includegraphics[width=\linewidth]{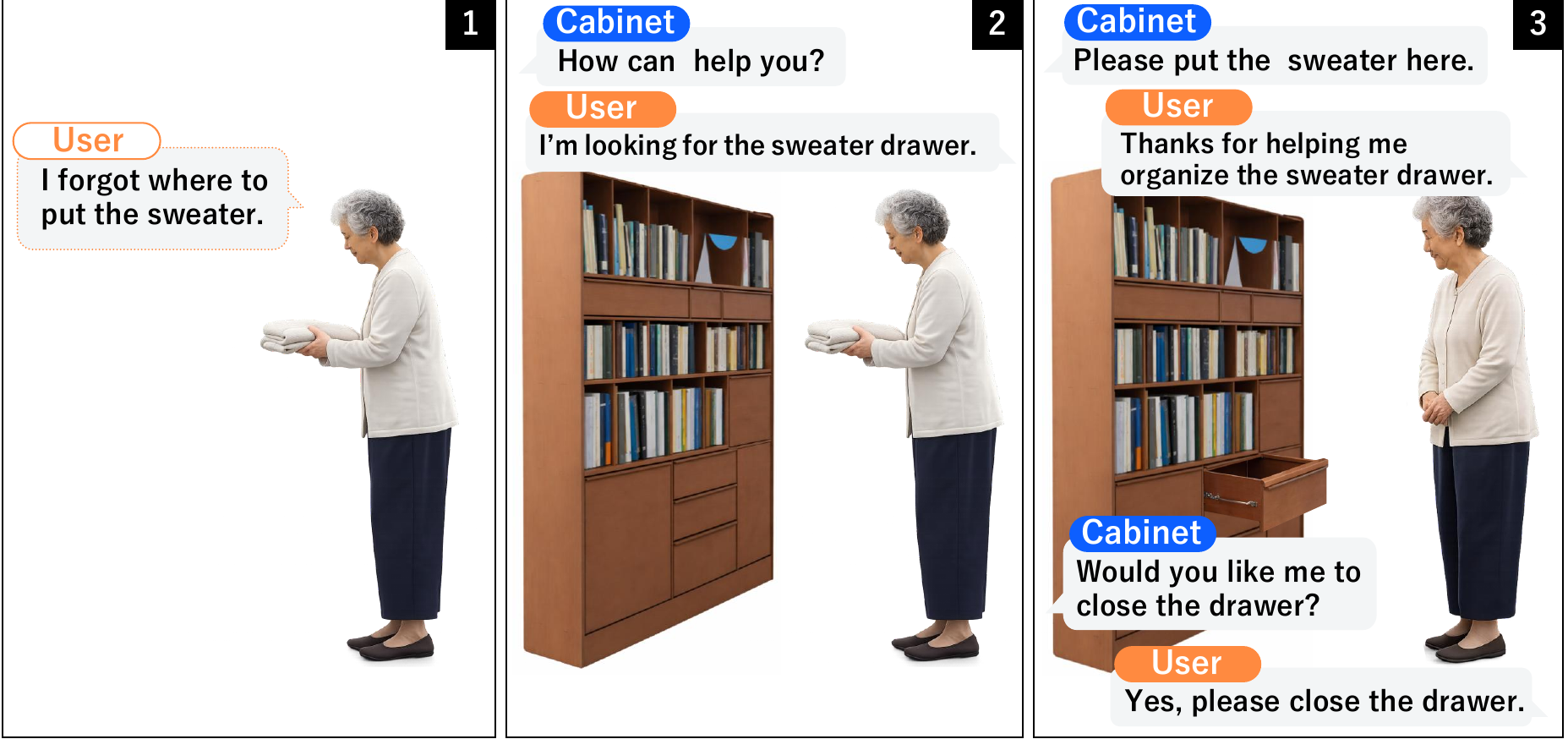}
        \Description{A storyboard illustrating a robot assisting a user in organizing household items.}
        \caption{Organizing an item scenario}
        \label{fig:app_1b}
    \end{subfigure}

    \begin{subfigure}{\linewidth}
        \centering
        \includegraphics[width=\linewidth]{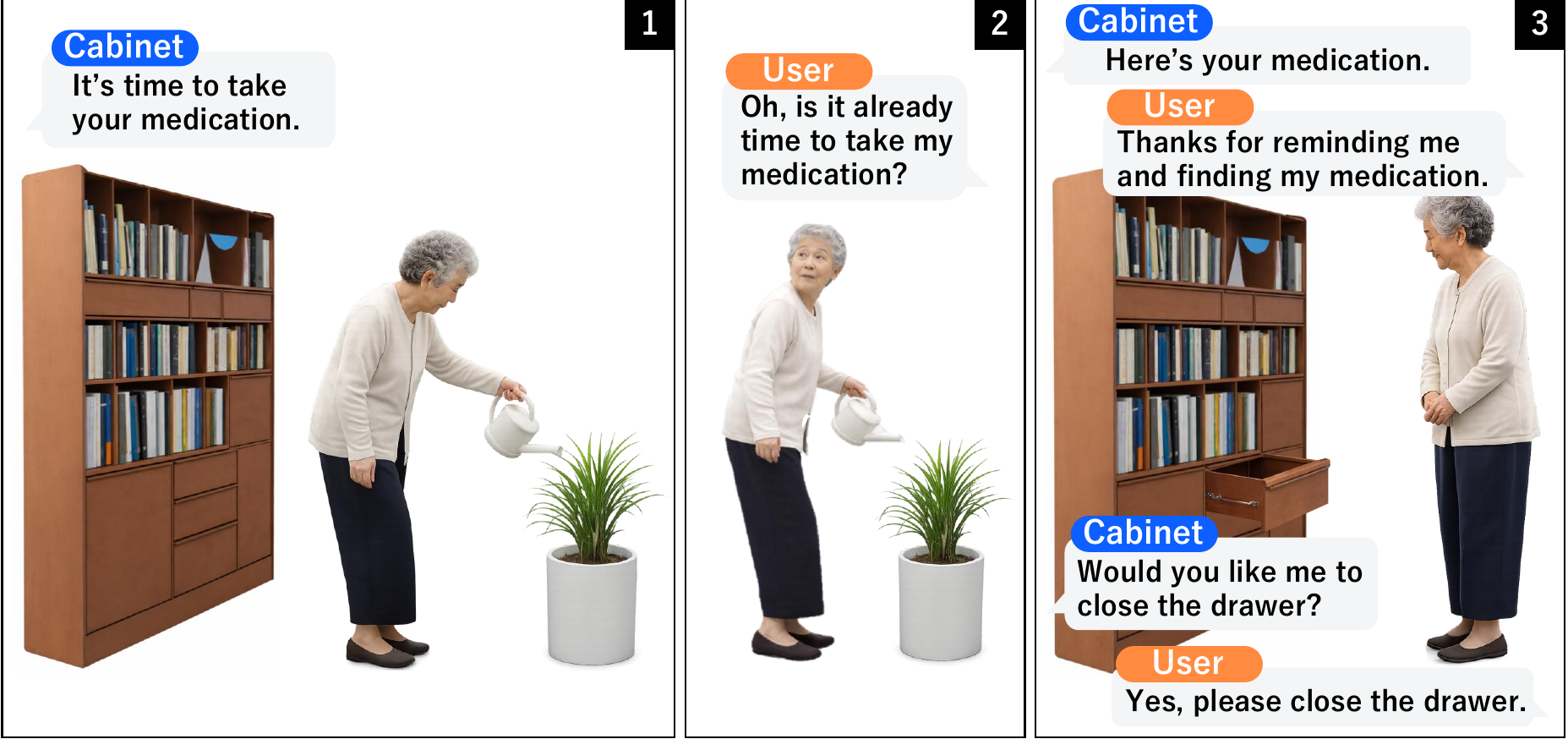}
        \Description{A storyboard showing the robot providing a voice reminder and opening a medicine drawer.}
        \caption{Medication reminder scenario.}
        \label{fig:app_1c}
    \end{subfigure}

    \begin{subfigure}{\linewidth}
        \centering
        \includegraphics[width=\linewidth]{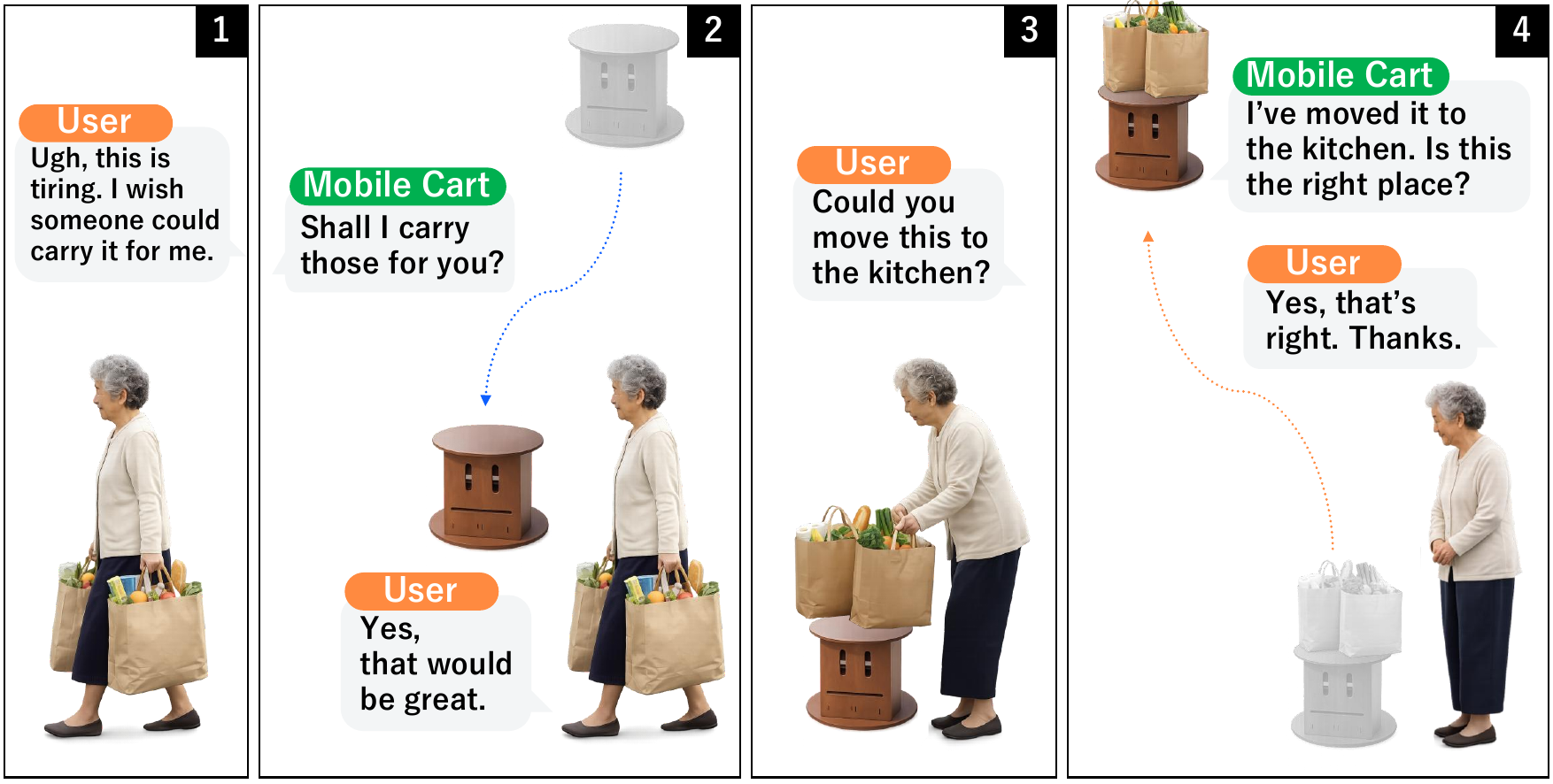} 
        \Description{A storyboard where the robot carries heavy objects to assist the user.}
        \caption{Transporting objects scenario.}
        \label{fig:app_1d}
    \end{subfigure}

%\par\vspace{-1em}
    \caption{Representative scenarios}
    \label{fig:appendix_scenarios_all} 

\end{figure}

We propose the following service scenarios utilizing a Cabinetbot and a Cart-like mobile platform to address the three recurrent activities and associated pain points identified in Section 2 (finding/organizing items, taking medication, and transporting objects). In these scenarios, the Cabinetbot provides item organization/finding and medication reminders, while the Cart-like mobile platform assists with transporting and placing objects. We illustrate four representative scenarios (S1–S4) as shown in Figure~\ref{fig:appendix_scenarios_all}.

\begin{itemize}[leftmargin=1.4em, labelsep=0.5em, itemsep=0.2em, topsep=2pt, partopsep=0pt]
  \item \textbf{S1 Finding an item} (Figure~\ref{fig:app_1a}): When a user asks for help finding a needed item, the agent assesses the situation (e.g., uncertainty about the item’s location), suggests likely drawer locations, and opens the selected drawer upon user confirmation. After retrieval, it confirms the follow-up action (e.g., closing the drawer), reducing surprise and unintended actuation.

  \item \textbf{S2 Organizing an item} (Figure~\ref{fig:app_1b}): When a user requests guidance during an organizing task, the agent recommends an appropriate storage location (drawer) and opens it if needed. After the item is placed, the system preserves user control through a confirmation step rather than automatically closing the drawer.

  \item \textbf{S3 Medication reminder} (Figure~\ref{fig:app_1c}): The agent provides time-based reminders and can explain the reason for the alert (e.g., scheduled dosage timing). After facilitating access to the medication, it confirms whether the dose was taken and whether the drawer should be closed, supporting completion of the routine.

  \item \textbf{S4 Transporting objects} (Figure~\ref{fig:app_1d}): When a user experiences physical difficulty or explicitly requests assistance, the agent proactively offers to use the mobile cart and, upon confirmation, performs the transport. Upon arrival, it verifies the final placement to ensure safety and predictability during movement.

\end{itemize}

\subsection{Conceptual Technical Operation}
Our proposed extension is a design sketch intended to support discussion of ``plausible implementation,''
and includes the following elements:

\begin{itemize}[leftmargin=1.4em, labelsep=0.5em, itemsep=0.2em, topsep=2pt, partopsep=0pt]
  \item \textbf{Context sensing:} Estimate only minimal state information---such as user presence/proximity,
  drawer states (open/closed), and cart location---to avoid excessive reliance on sensing.

  \item \textbf{Intent interpretation and clarification:} When user requests are ambiguous (e.g., ``Bring me that''),
  ask clarifying questions to reduce misinterpretation and risk.

  \item \textbf{Safety-constrained planning:} Adopt conservative defaults such as speed limits, obstacle avoidance,
  and confirmation gates before drawer closing or movement.

  \item \textbf{User-centred transparency/explainability:} Provide brief, comprehensible explanations of what the
  system will do and why (via speech and visual signals) to reduce uncertainty.

  \item \textbf{Fallback interaction and failure handling:} Ensure accessibility and equity by enabling alternative
  inputs (e.g., app buttons) so that functions remain usable even when speech/vision/LLM components fail.
\end{itemize}

In summary, in older adult wellbeing robotics, extending systems with generative AI and computer vision should not be framed as making ``smarter automation,'' but rather as enabling safer and more equitable interaction through confirmation, predictability, transparency, and fallback mechanisms.

\section{Video-Stimulated Follow-Up Interviews on the Redesigned Functions}

\subsection{Method}
To refine the proposed functions and surface interaction preferences, we conducted a second round of semi-structured interviews with the same four participants (P1–P4). We used video stimuli illustrating three core functions: automatic drawer opening/closing, equity-oriented cognitive reminders, and obstacle-avoiding navigation for goal-directed transport. The procedure had three steps: (1) showing short videos to establish a shared understanding; (2) explaining the intended purpose and mechanism; and (3) presenting activity-grounded scenarios (finding/organising items, medication intake, transporting objects) and probing appropriateness, preferred triggers, and improvement needs. Table~\ref{tab:Methods_second round interview}  summarises scenario-based interview probes for three functions.

% Table 2 위치 %
%%=======================================TABLE2===================================

% \begin{table}[H]
\begin{table}[!htbp]% 표 위치 고정 해제
\centering
\caption{Scenario-based interview probes for three functions} 
\label{tab:Methods_second round interview} 
\vspace{-6pt}
\footnotesize
\setlength{\tabcolsep}{4pt} 
\renewcommand{\arraystretch}{1.3} 

\newcolumntype{L}[1]{>{\RaggedRight\arraybackslash}m{#1}}
\newcolumntype{C}[1]{>{\centering\arraybackslash}m{#1}}

\setlist[itemize]{
    leftmargin=1.2em, 
    nosep, 
    topsep=3pt, 
    before=\vspace{0.1\baselineskip}, 
    after=\vspace{-0.1\baselineskip}
}

\begin{tabular}{C{2.2cm} L{5.5cm}}
\toprule 
\textbf{Function} & \centering \textbf{Scenario-based probes} \tabularnewline 
\midrule 

% 첫 번째 행
Automatic drawer opening/closing & 
\begin{itemize}
    \item Would this help when finding or organising items?
    \item How should the drawer be triggered?
    \item In what other situations could this be used?
    \item What additional functions are needed?
\end{itemize} \\ \addlinespace[1em]

% 두 번째 행
Equity-oriented cognitive reminders & 
\begin{itemize}
    \item Would this help when taking medication or routine situations?
    \item How would you like to be notified?
    \item In what other situations could this be used?
    \item What additional functions are needed?
\end{itemize} \\ \addlinespace[1em]

% 세 번째 행
Obstacle-avoiding navigation for goal-directed transport & 
\begin{itemize}
    \item Would this help when carrying heavy or multiple items?
    \item How should transport be initiated?
    \item In what other situations could this be used?
    \item What additional functions are needed?
\end{itemize} \\ 
\bottomrule 
\end{tabular}
\end{table}

%\vspace{-10pt}

\subsection{Findings}

Participants tended to reason less about \textbf{standalone functional capabilities} and more about \textbf{service choreography}, when the system should initiate, how it should ask, and what confirmations are needed. Table~\ref{tab:findings_summary} summarises the categorised responses. Three cross-cutting patterns emerged:

\begin{itemize}[leftmargin=1.4em, labelsep=0.5em, itemsep=0.2em, topsep=2pt, partopsep=0pt]
    \item \textbf{(F1) Confirmation and Consent before Actuation:} Participants frequently requested explicit confirmation before closing drawers and before moving, especially to prevent surprise and physical risk.
    \item \textbf{(F2) Predictability and Legibility:} Participants wanted the system to state what it will do and why (e.g., where it will move), highlighting \textit{transparency} as a safety and trust mechanism.
    \item \textbf{(F3) Context-sensitive Autonomy:} Participants welcomed proactive help when their hands are full or during routine tasks (e.g., medication), but preferred reactive help in response to user's explicit requests in other situations, %(e.g., when the robotic unit transforms into a stationary seat),
    suggesting a need for \textbf{adjustable autonomy} and mode awareness.
\end{itemize}

%\vspace{-7pt}
% Table 3 위치 %
%%==========================================table3===============================

% \begin{table}[H]
\begin{table}[!htbp]% 표 위치 고정 해제
\centering
\caption{Second interview results} 
\label{tab:findings_summary}
\vspace{-6pt}
\footnotesize
\setlength{\tabcolsep}{2pt}
\renewcommand{\arraystretch}{1.4} 

\newcolumntype{L}[1]{>{\RaggedRight\arraybackslash}m{#1}}
\newcolumntype{C}[1]{>{\centering\arraybackslash}m{#1}}

\setlist[itemize]{
    leftmargin=0.9em, 
    nosep, 
    topsep=2pt, 
    before=\vspace{0.1\baselineskip}, 
    after=\vspace{-0.1\baselineskip}
}

\begin{tabular}{C{1.3cm} L{2.3cm} L{2.3cm} L{1.9cm}}
\toprule 
\textbf{Function} & \centering \textbf{Use Context / Situation} & \centering \textbf{Preferred Interaction} & \centering \textbf{Additional needs} \tabularnewline 
\midrule 

% 첫 번째 행
Automatic drawer opening /closing & 
\begin{itemize}
    \item Helpful when repeatedly opening and closing drawers due to forgotten locations.(P1, P2)
    \item Organising contexts requiring frequent drawer opens(P3, P4)
    \item Supports routines requiring attention and memory.(P3)
\end{itemize} & 
\begin{itemize}
    \item System-initiated voice prompts when finding items.(P1, P3)
    \item User-triggered opening via short voice commands.(P2)
    \item Automatic opening of the relevant drawer when an item is specified(P4).
\end{itemize} & 
 User-adjustable opening and closing speed to avoid being startled.(P1)
 \\ \addlinespace[1em]

% 두 번째 행
Equity-oriented cognitive reminders & 
\begin{itemize}
    \item Useful for remembering medication and daily routines in time-sensitive situations.(P1--P4)
    \item Helpful for recalling what to bring before going out.(P3)
\end{itemize} & 
\begin{itemize}
    \item Voice reminders with medicine-drawer opening at medication time.(P1, P2)
    \item Reminders that include context-specific explanations.(P4)
\end{itemize} & 
 Voice and LED cues for recalling key items before leaving.(P3)
 \\ \addlinespace[1em]

% 세 번째 행
Obstacle-avoiding navigation for goal-directed transport & 
\begin{itemize}
    \item Useful when carrying heavy or multiple items.(P1--P4)
    \item Helpful when the robot comes to the entrance so I can sit to put on shoes.(P1)
\end{itemize} & 
\begin{itemize}
    \item Automatic Approach when both hands are full with items.(P1--P4)
    \item Approach as a chair only when explicitly requested.(P1)
\end{itemize} & 
 Advance verbal notice of how it will move indoors.(P2)
 \\ 
\bottomrule 
\end{tabular}
\end{table}
%%==================================================================================

\section{Equity-Centred Design Implications and Open Questions}
Grounded in the interviews, we outline design implications that align with themes of \textbf{ethics, justice, and responsible development} for equitable robotics for wellbeing.

\begin{itemize}[leftmargin=1.4em, labelsep=0.5em, itemsep=0.2em, topsep=2pt, partopsep=0pt]
    \item \textbf{D1. Preserve Autonomy (Adjustable Autonomy):} Provide user-configurable levels of initiative (user vs. system-initiated) and ensure actions are easy to pause, cancel, or override.
    \item \textbf{D2. Legible System State and Intent:} Use brief, user-centered explanations and \textit{multimodal cues} (voice, lights, motion previews) to reduce uncertainty and anxiety.
    \item \textbf{D3. Safety and Low-surprise Actuation:} Gate risky actions (e.g., movement, closing drawers) behind confirmations and adopt conservative default settings.
    \item \textbf{D4. Recognition over Recall:} Externalize storage memory (what/where) and medication status (taken/not) to reduce cognitive burden without requiring sustained attention.
    \item \textbf{D5. Fairness in Sensing and Language:}  Evaluate detection errors across diverse bodies (e.g., mobility aids), speech patterns, and lighting and provide fallback input channels.
    \item \textbf{D6. Privacy as Wellbeing Requirement:} Prefer on-device processing and ensure data practices are transparent, minimal, and revocable.
    \item \textbf{D7. Sustainability and Deployment:} Embed capabilities into familiar furniture to avoid stigmatizing aesthetics and consider cost models for equitable access.
\end{itemize}

% D7 불릿 바로 밑에 배치하기 위한 설정
%\vspace{4pt} 
%\noindent \textbf{Workshop Questions:} \textit{How should we operationalize `equity' for robotic furnishings? What evaluation protocols capture differential error burdens and long-term adoption? How can co-design be scaled to include caregivers and clinicians while respecting privacy?}

\section{Conclusion}
This position paper connects ADL/IADL-grounded pain points of older adults with an equity-centred concept for LLM-augmented robotic furnishing-agent. By combining early needs assessment with video-stimulated elicitation of interaction preferences, we highlight concrete requirements, confirmation, predictability, adjustable autonomy, and multimodal feedback that serve both usability and justice-oriented goals. We hope the workshop will help sharpen methods and metrics for equitable robotics for wellbeing, and inform responsible pathways from concept to deployment.

\begin{acks}
This work was supported by the Sejong Science Fellowship (RS-2025-00559234), funded by the National Research Foundation (NRF); the KIST Institutional Program (26E0211); the Technology Innovation Program (RS-2024-00419883), funded by the MOTIE; and the WISET (Grant No. 2025-560), funded by the Ministry of Science and ICT (MSIT).
\end{acks}

\bibliographystyle{ACM-Reference-Format}
\bibliography{sample-base}

\appendix

%%
%% If your work has an appendix, this is the place to put it.

\end{document}